\documentclass[3p,times,procedia]{elsarticle}
\usepackage[bookmarks=false]{hyperref}
    \hypersetup{colorlinks,
      linkcolor=blue,
      citecolor=blue,
      urlcolor=blue}

\usepackage{amsmath,amsfonts,amssymb}
\usepackage{algorithmic}
\usepackage{algorithm}
\usepackage{array}
\usepackage{url}
\usepackage{graphicx}
\usepackage{placeins}
\usepackage[utf8]{inputenc}
\usepackage{amsthm}
\usepackage{booktabs}
\usepackage{bbding}
\usepackage[dvipsnames]{xcolor}
\usepackage[normalem]{ulem}
\usepackage{subcaption}

\newcommand{\best}[1]{{\boldmath\color{green!55!black}#1}}
\newcommand{\sndfm}[1]{{\color{blue}\uline{#1}}}

\definecolor{commentGreen}{rgb}{0,0.5,0.05}

\newcommand{\safeincludegraphics}[3][]{%
    \IfFileExists{#2}{\includegraphics[width=\linewidth,#1]{#2}}{%
        \fbox{\parbox[c][2.8cm][c]{0.9\linewidth}{\centering #3\\[2pt]\texttt{#2}}}%
    }%
}

\usepackage{amssymb}

\usepackage[figuresright]{rotating}

\usepackage{xspace}
\newcommand{\etal}{\textit{et al.}\xspace}

\begin{document}
\begin{frontmatter}




\title{Few-Medoids: An Embarrassingly Simple Coreset Selection Method for
Few-Shot Knowledge Distillation}


\author[a]{Cemil-Andrei Dilmac} 
\author[a]{Florinel-Alin Croitoru}
\author[a]{Radu Tudor Ionescu\corref{cor1}}

\address[a]{Department of Computer Science, University of Bucharest, 15G Iuliu Maniu, Bucharest 061075, Romania}

\begin{abstract}
Coreset selection aims to identify a small and highly representative subset of a massive dataset for efficient model training. The problem remains challenging even in the few-shot knowledge distillation (KD) setup, where a full-scale pre-trained teacher informs the student network. Typical sample selection strategies often struggle to surpass the random selection baseline. In this paper, we showcase few-medoids, an embarrassingly simple coreset selection strategy that chooses the samples closest to the centroid (average image) of each class. We present extensive KD experiments on four datasets, covering a wide range of image classification problems, and three teacher-student model pairs, comprising both convolutional and transformer networks. Although the proposed method is embarrassingly simple, our empirical results indicate that few-medoids is able to consistently surpass the random selection baseline, as well as the other coreset selection strategies. We therefore consider that few-medoids can be used as a drop-in replacement for commonly-used baselines (e.g.~herding or k-center Greedy), in future research on coreset selection. To reproduce the reported results, we publicly release our code at \url{https://github.com/CemilAndreiDilmac/Few-Shot-KD-Coreset}. 
\end{abstract}

\cortext[cor1]{Corresponding author.}

\end{frontmatter}

\setlength{\abovedisplayskip}{3.8pt}
\setlength{\belowdisplayskip}{3.8pt}
\setlength{\abovedisplayshortskip}{3pt}
\setlength{\belowdisplayshortskip}{3pt}

\email{radu.ionescu@fmi.unibuc.ro}


\section{Introduction}

The advancements in modern deep learning have revolutionized various domains, from computer vision~\cite{Croitoru-TPAMI-2023, Dosovitskiy-ICLR-2021, He-CVPR-2016, Liu-ICCV-2021, Rombach-CVPR-2022} to natural language processing~\cite{ Rafailov-NeurIPS-2023, Touvron-arXiv-2023, Vaswani-NeurIPS-2017}. However, this success is often obtained by a paradigm that prioritizes performance metrics over computational efficiency. State-of-the-art models increasingly rely on massive and potentially redundant datasets, as well as over-parameterized architectures that lead to extremely high memory and computational costs during both training and deployment phases. As the environmental and economic impacts of these computational demands become more pronounced~\cite{Faiz-ICLR-2024, Li-ACM-2025, Schwartz-ACM-2020}, developing sustainable and resource-efficient practices has emerged as an important requirement for the deep learning community~\cite{Schwartz-ACM-2020}.

To address the severe inefficiencies associated with model training, two equivalent effective solutions that researchers have explored are coreset selection and data pruning~\cite{Agarwal-2005, Dai-ECCV-2020, Hao-CVPR-2025, Lee-CVPR-2024, Moser-arXiv-2026}. These methodologies seek to transform a massive dataset into a minimal, highly representative subset that preserves the statistical and structural distribution of the original data. Usually, coreset selection methods~\cite{Baruch-ICML-2025, Dai-ECCV-2020, Hao-CVPR-2025, He-CVPR-2024, Lee-CVPR-2024} leverage statistical properties to identify and discard uninformative or redundant examples, while retaining only the most critical samples. Consequently, coreset selection allows models to be trained on highly compact datasets, reducing computational overhead and facilitating effective few-shot model training with reduced performance impact. 

Complementary to the training optimizations offered by coreset selection, knowledge distillation (KD)~\cite{Baruch-ICML-2025, Fukuda-Interspeech-2017, Hinton-arXiv-2015,  Zhao-CVPR-2022} targets resource efficiency at the inference stage. This technique works by compressing the predictive capabilities and learned feature representations of a large teacher model into a much smaller and more resource-efficient student network. The main goal is to ensure low latency and to reduce memory usage during real-world deployment. KD has been extensively explored, as deploying heavy models on resource-constrained edge devices remains a critical industry bottleneck \cite{Grigore-WACV-2025,Tsuyuki-Arvix-2026}. Although coreset selection and knowledge distillation share the objective of resource optimization, they have largely been treated as isolated domains. The former is applied exclusively for training optimization, while the latter is reserved for inference optimization. Consequently, there is a notable absence of literature investigating the intersection of these two methodologies. Specifically, to the best of our knowledge, there are only a few recent works \cite{Baruch-ICML-2025,Chen-ICLR-2025} that investigated the application of coreset selection to knowledge distillation, showcasing that the random selection baseline often outperforms several other specialized methods. Overall, utilizing a coreset for distillation is underexplored, despite the benefits that are brought by this setup. Specifically, it could simultaneously reduce the computational burden of training the student model, while ensuring the student still learns effectively from the teacher's representations.

In this work, we address the identified research gap and expand the exploration of coreset selection in conjunction with KD. Moreover, we introduce a novel, simple, yet powerful baseline called \emph{few-medoids}, designed to optimally identify representative samples for distillation by leveraging the latent space of the teacher. Specifically, few-medoids analyzes the per-class feature representations extracted from the teacher. Rather than relying on complex heuristics, it determines the most informative samples within a given class by calculating their centrality. Specifically, we select the data points that exhibit the least average distance to all other samples in the same class. Through comprehensive empirical validation of various convolutional and transformer architectures on four image datasets, we demonstrate that by capturing these geometric cluster centers, few-medoids can consistently outperform previously established baselines across a diverse range of experimental setups.

In summary, our contribution is three-fold:
\begin{itemize}
    \item We explore the effectiveness of adapting several existing baseline coreset methods to the distillation pipeline.
    \item We introduce \emph{few-medoids}, a simple and powerful coreset selection method that leverages the latent space of the teacher model.
    \item We evaluate our proposed method against established baselines across four datasets and multiple model architectures.
\end{itemize}

\section{Related Work}

Recent advances in deep learning have been largely driven by scale, with improved performance often achieved by training continuously growing models on increasingly larger datasets \cite{Dosovitskiy-ICLR-2021, Enevoldsen-ICLR-2025, He-CVPR-2016, Liu-ICCV-2021, Lupacscu-IF-2026, Muennighoff-ACL-2023, Rafailov-NeurIPS-2023, Rombach-CVPR-2022, Touvron-arXiv-2023, Vaswani-NeurIPS-2017}. However, this trend comes at a substantial computational cost, as state-of-the-art performance typically requires significant training resources \cite{Schwartz-ACM-2020}. Moreover, supervised learning relies on large amounts of annotated data, and acquiring such labels is often expensive, time-consuming, and, in many domains, requires expert knowledge. These limitations have motivated research directions that seek to reduce either the amount of training data or the labeling burden, while maintaining competitive performance \cite{Agarwal-2005, Dai-ECCV-2020, Grigore-WACV-2025, Hao-CVPR-2025, Lee-CVPR-2024, Moser-arXiv-2026}.

One important line of research is active learning \cite{Ducoffe-arXiv-2018, Hondru-AIR-2025, Hsu-AAAI-2015, Parvaneh-CVPR-2022, Sener-arXiv-2017}, which aims to reduce annotation cost by selecting the most informative samples from an unlabeled pool for labeling. Coreset selection \cite{Agarwal-2005,Griffin-WACV-2026, Guo-DEXA-2022, Hao-CVPR-2025, Lee-CVPR-2024, Moser-arXiv-2026, Zheng-ICLR-2025} and data pruning~\cite{Ayed-TMLR-2023,Baruch-ICML-2025, Dai-ECCV-2020, He-CVPR-2024, Sorscher-NeurIPS-2022,Vysogorets-ICLR-2025, Yang-ICLR-2023} are closely related, as they also seek to identify a compact yet informative subset of examples. Coreset selection has its origins in computational geometry, where early work~\cite{Agarwal-2005} studied small subsets that approximate the original set with respect to a given objective. The three paradigms are connected by their shared objective of selecting samples that are especially useful for learning, often according to criteria such as informativeness, representativeness, or diversity~\cite{ Baruch-ICML-2025, Dai-ECCV-2020, Guo-DEXA-2022, Hao-CVPR-2025,He-CVPR-2024, Lee-CVPR-2024, Sener-arXiv-2017,   Sorscher-NeurIPS-2022}. The main distinction lies in the learning setting: active learning operates on unlabeled data and focuses on which samples should be annotated, whereas coreset selection and data pruning typically assumes that labels are already available, and instead, aims to construct a smaller subset that can effectively approximate the full labeled training set.

Following the taxonomy proposed by Moser~\etal~\cite{Moser-arXiv-2026}, coreset selection methods can be broadly grouped into three categories: training-based, training-free, and label-free approaches. Training-based methods rely on model-dependent signals observed during optimization, such as gradients~\cite{Everaert-ICLR-2024, Paul-NeurIPS-2021}, forgetting events~\cite{Toneva-arXiv-2018}, decision-boundary information~\cite{Ducoffe-arXiv-2018}, gradient matching~\cite{Mirzasoleiman-ICML-2020}, and bilevel optimization~\cite{Killamsetty-AAAI-2021}. While often effective, these methods usually require repeated optimization, retraining, or additional scoring procedures, which introduces substantial computational overhead. In contrast, training-free methods rely on criteria that are independent of training dynamics, making them especially attractive in large-scale settings where efficiency and scalability are critical. This category includes simple strategies such as random sampling, as well as geometry-based methods, which remain competitive and serve as strong reference points not only in the coreset selection literature~\cite{Guo-DEXA-2022,  Lee-CVPR-2024, Moser-arXiv-2026, Zheng-ICLR-2023}, but also in dataset pruning~\cite{He-CVPR-2024, Maharana-ICLR-2024, Sener-arXiv-2017,   Sorscher-NeurIPS-2022, Xia-ICLR-2023}, as both directions pursue the reduction of the training set, while preserving model performance. A third category, label-free methods, aims to estimate sample importance without using ground-truth labels, often through clustering or pre-trained models~\cite{Griffin-WACV-2026, Tan-ICLR-2025, Zheng-ICLR-2025}.

Our work belongs to the training-free category, and more specifically to geometry-based coreset selection~\cite{Chen-UAI-2010, Guo-DEXA-2022,Moser-arXiv-2026, Rebuffi-CVPR-2017, Welling-ICML-2009}. Accordingly, the most relevant baselines for comparison are random selection, herding, and k-center Greedy. Random selection provides a simple baseline with negligible computational overhead, but it does not explicitly enforce representativeness or diversity, capturing dataset structure only through uniform sampling. In contrast, the two geometric baselines rely on the assumption that samples that are close in the latent space tend to share similar properties, so the geometry of that space carries meaningful information about the data distribution. Herding is designed to promote representativeness by greedily selecting samples such that the mean of the selected subset approximates the mean of the full dataset~\cite{Chen-UAI-2010,Rebuffi-CVPR-2017, Welling-ICML-2009}. K-center Greedy instead emphasizes diversity by selecting samples that maximize coverage of the data space under a geometric distance criterion~\cite{Sener-arXiv-2017}. 

In the context of data pruning for knowledge distillation (KD)~\cite{Fukuda-Interspeech-2017, Grigore-WACV-2025, Hinton-arXiv-2015, Zhao-CVPR-2022}, recent work~\cite{Baruch-ICML-2025} has shown that the random sampling method is  a surprisingly strong baseline. Moreover, a recent pruning study shows that the examples preferred in standard supervised subset selection are not always the most suitable for distillation. In particular, very hard examples may be suboptimal because the student cannot reliably match the teacher on them, while medium-difficulty ones provide a better trade-off between informativeness and learnability, leading to smoother decision boundaries and more stable knowledge transfer~\cite{Chen-ICLR-2025}. Specifically, Chen \etal~\cite{Chen-ICLR-2025} rank data points by prediction difficulty using the teacher's cross-entropy loss on the ground-truth label, and select the medium-difficulty samples. Differently, few-medoids ranks samples within each class by their average Euclidean distance to all members of the same class, operating in the latent space of the teacher model, thereby favoring geometrically central points. As a result, the two methods retain different types of data.

\section{Method}

\noindent
\textbf{Problem statement.}
Coreset selection seeks to reduce a labeled dataset ${D}=\{(x_i, y_i)\}_{i=1}^n$ into a substantially smaller subset ${C}=\{(x_j, y_j)\}_{j=1}^p$, called \emph{coreset}, such that $C \subset D$ and $p \ll n$. The goal is to preserve the most representative samples from each class, such that the empirical risk of a model $h$ trained on $C$ is as close as possible to the empirical risk of $h$ trained on $D$.


\noindent
\textbf{Few-medoids.}
In our KD-based approach, the coreset selection process is executed independently for each class, using the representations extracted by a teacher model. We denote this latent feature mapping of the teacher as $f_\phi:\mathcal{X}\rightarrow \mathbb{R}^d$, where $\mathcal{X}$ is the input space, $\phi$ denotes the parameters of the teacher, and $d$ represents the feature space dimension.

Let $X_c = \{x_i \mid  y_i=c\}_{i=1}^m$ denote the set of training samples belonging to a given class $c$. Few-medoids assigns a score $s_i$ to each sample, which is equal to its average Euclidean ($L_2$) distance to all other elements in the class. These values are then ranked in ascending order, so that lower scores indicate greater representativeness. The top-ranked image is therefore the class medoid, while the remaining ones can be viewed as progressively weaker medoid-like representatives under the same centrality criterion. In Algorithm~\ref{alg:medoid_ranking}, we summarize the proposed ranking method. In line 2, we initialize the list of importance scores $S_{\!c}$. Subsequently, in lines 3-5, we determine these scores by calculating the distances between the latent feature vector $f_\phi(x_i)$ of each sample and the latent feature vectors of all samples in class $c$. Lastly, we sort the samples in $X_c$ according to $S_{\!c}$ and store the top $k$ elements in $C_c$. Then, the final corset $C$ is given by:
\begin{equation}
C = \bigcup_{j=1}^l C_j, 
\end{equation}
where $l$ is the number of classes.



\begin{algorithm}[t]
\caption{Few-medoids}
\label{alg:medoid_ranking}
\begin{algorithmic}[1]
\STATE \textbf{input} $f_\phi$ - teacher model, $X_c = \{x_i \mid  y_i=c\}_{i=1}^m$ - training examples belonging to class $c$, $k$ - the number of returned samples.
\STATE $S_{\!c} \leftarrow \emptyset$; \textcolor{commentGreen}{$\lhd$ initialize $S_{\!c}$ as an empty set}
\FOR{$i = 1, \dots, m$}
    \STATE $s_i \leftarrow \frac{1}{m} \sum_{j=1}^{m} \|f_\phi(x_i) - f_\phi(x_j)\|_2$; \textcolor{commentGreen}{$\lhd$ compute average distance between $x_i$ and all data points in class $c$}
    \STATE $S_{\!c} \leftarrow S_{\!c} \cup \{ s_i \}$; \textcolor{commentGreen}{$\lhd$ add average distance $s_i$ to the set $S_{\!c}$}
    
\ENDFOR
\STATE $I \leftarrow \mathbf{argsort}(S_{\!c}, \texttt{`asc'})$; \textcolor{commentGreen}{$\lhd$ sort samples in $S_{\!c}$ in ascending order and return the sorted indexes}
\STATE $C_c \leftarrow X_c[I[0:k]]$; \textcolor{commentGreen}{$\lhd$ take the top $k$ examples in $X_c[I]$ and store them in $C_c$}
\STATE \textbf{output} $C_c$
\end{algorithmic}
\end{algorithm}

\noindent
\textbf{Knowledge distillation.}
For our knowledge distillation framework, we adopt the soft-label objective introduced by Hinton et al.~\cite{Hinton-arXiv-2015}. For a given input batch $X$, let $z_\theta(X)$ and $z_\phi(X)$ denote the output logits produced by the student and teacher models, respectively. We define the distillation loss, as follows:
\begin{equation}\label{eq_L_KD}
    \mathcal{L}_{\text{KD}}(X,\phi, \theta) = \tau^2 \cdot D_{\text{KL}}\left(p_\phi(X) \,\|\, p_\theta(X)\right),
\end{equation}
where $\tau$ is a temperature parameter, $D_\mathrm{KL}$ is the Kullback–Leibler (KL) divergence, and $p_\phi$ and $p_\theta$ are soft probability distributions given by:
\begin{equation} 
p_\phi(X) = \text{softmax}\left(\frac{z_\phi(X)}{\tau}\right),
\qquad
p_\theta(X) = \text{softmax}\left(\frac{z_\theta(X)}{\tau}\right).
\end{equation}

The soft-target loss in Eq.~\eqref{eq_L_KD} is combined with the standard cross-entropy loss, denoted by $\mathcal{L}_{\text{CE}}$, computed using the ground-truth hard labels $Y$. The total loss optimized by the student network is thus defined as:
\begin{equation}
    \mathcal{L}_{\text{total}} = \lambda_{1}\cdot\mathcal{L}_{\text{KD}}(X,\phi, \theta) + \lambda_2 \cdot \mathcal{L}_{\text{CE}}(X, Y, \theta),
\end{equation}
where $\lambda_{1}$ and $\lambda_{2}$ are hyperparameters weighting the two objectives.

\section{Experiments}
\noindent
\textbf{Datasets. }We conduct experiments on four benchmark datasets: CIFAR-10 \cite{Krizhevsky-2009}, CIFAR-100 \cite{Krizhevsky-2009}, Oxford Flowers 102 \cite{Nilsback-ICVGIP-2008}, and Food-101 \cite{Bossard-ECCV-2014}. For CIFAR-10, CIFAR-100, and Food-101, the validation set was created from the official training split using a stratified 80/20 split to preserve class balance, whereas for Oxford Flowers 102, we used the official split. CIFAR-10 and CIFAR-100 have a fixed image resolution of $32 \times 32$ pixels, while Oxford Flowers 102 and Food-101 contain images with variable resolutions. The classes in CIFAR-10 represent general object categories, primarily animals and vehicles. CIFAR-100 spans a broader range of categories, including animals, plants, household objects, people, and vehicles. In Oxford Flowers 102, the classes correspond to flower species, while in Food-101, they represent types of food dishes. Detailed statistics for all datasets are presented in Table~\ref{tab:datasets}.

\begin{table}[t!]
\caption{Statistics across object class recognition datasets. We report the number of classes, as well as the number of images and the number of images per class for each split: training, validation and test.}
\label{tab:datasets}
\vspace{-0.3cm}
\begin{tabular*}{\hsize}{@{\extracolsep{\fill}}lccccccc@{}}
\toprule
 & & \multicolumn{3}{c}{Total images} & \multicolumn{3}{c}{Images per class} \\
\cmidrule(lr){3-5} \cmidrule(lr){6-8}
Dataset & Classes & Train & Val & Test & Train & Val & Test \\
\midrule
CIFAR-10   & 10  & 40,000 & 10,000 & 10,000 & 4,000 & 1,000 & 1,000 \\
CIFAR-100  & 100 & 40,000 & 10,000 & 10,000 & 400   & 100   & 100   \\
Oxford Flowers 102 & 102 & 1,020  & 1,020  & 6,149  & 10    & 10    & 20-238 \\
Food-101    & 101 & 60,600 & 15,150 & 25,250 & 600   & 150   & 250   \\
\bottomrule
\end{tabular*}
\vspace{-0.5cm}
\end{table}


\noindent
\textbf{Underlying neural models.}
We evaluate several teacher-student combinations, including pairs based on the same model family, as well as a cross-architecture setting. The former pairs include ResNet-34$\rightarrow$ResNet-18 \cite{He-CVPR-2016} and ViT-B/16$\rightarrow$ViT-Small \cite{Dosovitskiy-ICLR-2021}, while the latter is represented by ViT-B/16$\rightarrow$ResNet-50. In all cases, the teacher is initialized with ImageNet-1K pre-trained weights and then fine-tuned on the target dataset. The teacher is used both to provide the distillation signal and to extract features during the sample selection stage.

Each student is also trained independently on the full training set using standard supervised learning, providing a reference for the upper performance bound, outside the few-shot distillation framework. In this setting, ResNet-18 and ResNet-50 are trained from scratch, whereas ViT-Small is initialized from pre-trained weights. These models serve as supervised baselines. In the proposed framework, KD is performed only on the subsets selected by coreset selection methods, enabling a direct comparison between supervised training on the full training set and distillation on the selected subsets. The ViT-B/16$\rightarrow$ViT-Small pair also allows evaluation in a pre-trained teacher-student setting.

\noindent
\textbf{Hyperparameters.}
All architectures are trained for 100 epochs using a Cosine Annealing learning rate scheduler. ResNet-based models are optimized via SGD with momentum $0.9$, while vision transformers are optimized via AdamW. A batch size of $128$ is used for teacher training, while student models are trained with a batch size of $32$. The learning rate varies between $5\cdot 10^{-2}$ and $10^{-4}$. Students trained via KD on the selected subsets use the same optimization settings, training schedule, and data augmentation pipeline as the corresponding architectures trained on the full training set. The distillation temperature is set to $\tau = 4$. The loss weights are set to $\lambda_1=0.6$ and $\lambda_2=0.4$.



\begin{table}[t!]
\caption{Test accuracy rates (\%) of the five architectures trained on the full training set of each dataset. The report accuracy rates serve as empirical upper bounds for the few-shot KD experiments.}
\label{tab:supervised_baselines}
\vspace{-0.3cm}
\begin{tabular*}{\hsize}{@{\extracolsep{\fill}}lcccc@{}}
\toprule
Model & 
CIFAR-10 & CIFAR-100 & Oxford Flowers 102 & Food-101 \\
\midrule
ResNet-18  & 
94.03 & 74.88 & 49.58 & 81.32 \\
ResNet-34  & 
96.94 & 83.16 & 90.82 & 84.47 \\
ResNet-50  & 
92.27 & 72.98 & 37.92 & 80.12 \\
ViT-Small & 
97.09 & 85.72 & 98.04 & 88.40 \\
ViT-B/16  & 
97.31 & 85.85 & 93.15 & 86.78 \\
\bottomrule
\end{tabular*}
\vspace{-0.5cm}
\end{table}

\noindent
\textbf{Baselines. }To benchmark the performance of \emph{few-medoids}, we compare it against four baseline selection strategies. For a fair comparison, we applied these baselines in the latent space of the teacher. The first is \emph{random selection}, where we uniformly sample $k$ instances per class without replacement. The second is \emph{herding}~\cite{Welling-ICML-2009}, which ensures that the running mean of the selected subset progressively approximates the true class centroid. The third is k-center Greedy~\cite{Sener-arXiv-2017}, an algorithm that iteratively selects samples to maximize diversity in the feature space. Lastly, we introduce a custom baseline, PCA-guided matching, which leverages Principal Component Analysis (PCA) applied to the transposed feature matrix of the class samples. This transposition yields principal components that represent linear combinations of the data samples rather than the features. We select the original samples closest to these principal components to obtain distinct data samples that explain independent variations and promote subset diversity.

\noindent
\textbf{Evaluation protocol.}
We perform few-shot KD experiments for $k \in \{1,2,4,8,16,32,64,128\}$ samples per class. For Oxford Flowers 102, the evaluation is limited to $k \in \{1,2,4,8\}$, since each class only contains 10 training samples. For each selection strategy and each value of $k$, five independent selections are generated using different seeds. Each resulting subset is used for a separate training run. For a given run, the checkpoint retained for final evaluation is the one that achieves the highest validation accuracy during training. When multiple checkpoints reach the same validation accuracy, the one with the lowest validation loss is selected. This checkpoint is then evaluated on the official held-out test set to obtain the final test accuracy. The reported accuracy rates represent average values over five runs.

\begin{table}[t]
\caption{Few-shot KD test accuracy (\%) on CIFAR-10, for $k \in \{1,2,4,8,16,32,64,128\}$. Average accuracy rates and standard deviations are computed over five runs with different seeds. \textbf{\textcolor{green!55!black}{Bold green}} indicates the best result among coreset selection methods for a given $k$ and a given neural model pair, while \textcolor{blue}{\uline{blue underline}} indicates the second-best result.}
\label{tab:results_cifar10}
\vspace{-0.3cm}
\resizebox{\linewidth}{!}{\footnotesize\begin{tabular}{lcccccccc}
\toprule
Method & $k{=}1$ & $k{=}2$ & $k{=}4$ & $k{=}8$ & $k{=}16$ & $k{=}32$ & $k{=}64$ & $k{=}128$ \\
\midrule
\multicolumn{9}{l}{{\it{ResNet-34 $\to$ ResNet-18}}} \\[2pt]
Few-medoids & \best{$18.05 \pm 1.23$} & \sndfm{$21.65 \pm 0.47$} & \sndfm{$22.02 \pm 0.55$} & \best{$30.16 \pm 0.56$} & \best{$34.31 \pm 1.00$} & \best{$41.68 \pm 0.43$} & \best{$47.01 \pm 0.66$} & \sndfm{$57.75 \pm 1.65$} \\
K-center    & $16.13 \pm 1.41$ & $16.71 \pm 0.62$ & $17.91 \pm 1.42$ & $20.10 \pm 0.51$ & $22.36 \pm 0.89$ & $25.30 \pm 1.54$ & $30.20 \pm 0.25$ & $36.88 \pm 0.67$ \\
Herding     & \sndfm{$17.17 \pm 1.46$} & \best{$22.25 \pm 0.26$} & \best{$22.89 \pm 1.43$} & \sndfm{$28.07 \pm 1.12$} & \sndfm{$31.73 \pm 0.78$} & \sndfm{$36.55 \pm 0.93$} & $43.55 \pm 0.30$ & $57.31 \pm 2.13$ \\
Random      & $16.13 \pm 1.41$ & $18.41 \pm 1.21$ & $19.51 \pm 1.60$ & $25.16 \pm 0.33$ & $29.98 \pm 0.63$ & $36.18 \pm 1.43$ & \sndfm{$44.61 \pm 1.82$} & \best{$58.91 \pm 2.61$} \\
PCA-guided matching    & $15.92 \pm 0.79$ & $15.86 \pm 1.06$ & $17.40 \pm 0.89$ & $20.50 \pm 0.77$ & $23.84 \pm 0.58$ & $28.41 \pm 0.73$ & $35.61 \pm 1.61$ & $52.56 \pm 1.18$ \\
\midrule
\multicolumn{9}{l}{{\it{ViT-B/16 $\to$ ResNet-50}}} \\[2pt]
Few-medoids & \sndfm{$14.68 \pm 2.66$} & \sndfm{$16.75 \pm 1.42$} & \sndfm{$14.51 \pm 2.34$} & $14.72 \pm 3.64$ & \best{$21.54 \pm 5.57$} & \best{$33.43 \pm 2.97$} & \best{$41.16 \pm 1.60$} & \best{$49.96 \pm 2.21$} \\
K-center    & $13.46 \pm 0.29$ & $13.41 \pm 1.14$ & $13.14 \pm 0.47$ & $12.77 \pm 1.55$ & $14.89 \pm 2.74$ & $14.79 \pm 4.10$ & $20.14 \pm 5.29$ & $28.14 \pm 5.84$ \\
Herding     & \best{$15.79 \pm 1.74$} & \best{$17.41 \pm 2.07$} & \best{$14.94 \pm 2.83$} & \best{$17.29 \pm 2.00$} & \sndfm{$20.43 \pm 3.60$} & $23.39 \pm 8.12$ & $28.62 \pm 10.18$ & $46.38 \pm 6.85$ \\
Random      & $13.46 \pm 0.29$ & $14.41 \pm 1.67$ & $12.45 \pm 1.52$ & \sndfm{$15.33 \pm 3.73$} & $18.13 \pm 3.23$ & \sndfm{$26.02 \pm 3.41$} & \sndfm{$29.67 \pm 5.04$} & \sndfm{$47.06 \pm 4.58$} \\
PCA-guided matching    & $12.84 \pm 0.94$ & $11.82 \pm 1.16$ & $11.70 \pm 0.75$ & $14.01 \pm 1.24$ & $19.28 \pm 2.91$ & $21.17 \pm 3.43$ & $28.32 \pm 2.86$ & $35.31 \pm 4.34$ \\
\midrule
\multicolumn{9}{l}{{\it{ViT-B/16 $\to$ ViT-Small}}} \\[2pt]
Few-medoids & \sndfm{$66.52 \pm 2.39$} & \best{$74.82 \pm 0.89$} & \sndfm{$78.93 \pm 1.55$} & $82.87 \pm 1.43$ & $84.32 \pm 0.44$ & $85.87 \pm 0.44$ & $86.94 \pm 1.27$ & $87.90 \pm 1.38$ \\
K-center    & $47.96 \pm 4.33$ & $49.99 \pm 3.95$ & $58.94 \pm 5.31$ & $69.69 \pm 4.23$ & $78.42 \pm 5.77$ & $87.48 \pm 1.67$ & $92.12 \pm 0.64$ & $93.21 \pm 0.29$ \\
Herding     & \best{$67.76 \pm 2.51$} & \sndfm{$73.62 \pm 3.40$} & \best{$83.55 \pm 1.06$} & \best{$88.64 \pm 1.32$} & \best{$91.04 \pm 1.16$} & \best{$93.59 \pm 0.20$} & \best{$94.53 \pm 0.60$} & \best{$94.46 \pm 0.24$} \\
Random      & $47.96 \pm 4.33$ & $58.39 \pm 2.91$ & $71.04 \pm 4.79$ & \sndfm{$83.57 \pm 1.98$} & \sndfm{$89.51 \pm 0.91$} & \sndfm{$92.38 \pm 0.76$} & \sndfm{$93.71 \pm 0.22$} & \sndfm{$94.17 \pm 0.74$} \\
PCA-guided matching    & $31.03 \pm 3.45$ & $44.40 \pm 3.46$ & $62.74 \pm 2.91$ & $71.16 \pm 2.73$ & $80.88 \pm 3.24$ & $87.04 \pm 0.91$ & $89.38 \pm 0.85$ & $90.90 \pm 0.66$ \\
\bottomrule
\end{tabular}}
\vspace{-0.3cm}
\end{table}



\begin{table}[t]
\caption{Few-shot KD test accuracy (\%) on CIFAR-100, for $k \in \{1,2,4,8,16,32,64,128\}$. Average accuracy rates and standard deviations are computed over five runs with different seeds. \textbf{\textcolor{green!55!black}{Bold green}} indicates the best result among coreset selection methods for a given $k$ and a given neural model pair, while \textcolor{blue}{\uline{blue underline}} indicates the second-best result.}
\label{tab:results_cifar100}
\vspace{-0.3cm}
\resizebox{\linewidth}{!}{\footnotesize\begin{tabular}{lcccccccc}
\toprule
Method & $k{=}1$ & $k{=}2$ & $k{=}4$ & $k{=}8$ & $k{=}16$ & $k{=}32$ & $k{=}64$ & $k{=}128$ \\
\midrule
\multicolumn{9}{l}{{\it{ResNet-34 $\to$ ResNet-18}}} \\[2pt]
Few-medoids & \best{$6.14 \pm 0.38$} & \best{$9.66 \pm 0.40$} & \best{$14.60 \pm 0.59$} & \best{$19.88 \pm 0.41$} & \best{$29.67 \pm 0.76$} & \sndfm{$41.99 \pm 0.92$} & $55.09 \pm 0.49$ & $63.83 \pm 0.20$ \\
K-center    & $4.31 \pm 0.40$ & $5.29 \pm 0.61$ & $6.41 \pm 0.45$ & $8.82 \pm 0.63$ & $13.39 \pm 0.15$ & $25.82 \pm 0.75$ & $45.57 \pm 0.89$ & $63.23 \pm 0.43$ \\
Herding     & \sndfm{$5.73 \pm 0.38$} & \sndfm{$8.49 \pm 0.23$} & \sndfm{$12.12 \pm 0.21$} & \sndfm{$16.93 \pm 0.27$} & \sndfm{$25.63 \pm 0.44$} & \best{$42.34 \pm 0.51$} & \best{$58.02 \pm 0.44$} & \best{$66.92 \pm 0.50$} \\
Random      & $4.31 \pm 0.40$ & $7.17 \pm 0.43$ & $10.28 \pm 0.31$ & $15.13 \pm 0.81$ & $24.66 \pm 0.40$ & $40.27 \pm 1.42$ & \sndfm{$56.40 \pm 0.33$} & \sndfm{$66.16 \pm 0.25$} \\
PCA-guided matching    & $3.47 \pm 0.18$ & $6.03 \pm 0.16$ & $8.18 \pm 0.19$ & $13.15 \pm 0.35$ & $23.65 \pm 0.68$ & $39.75 \pm 0.43$ & $55.08 \pm 0.32$ & $66.08 \pm 0.28$ \\
\midrule
\multicolumn{9}{l}{{\it{ViT-B/16 $\to$ ResNet-50}}} \\[2pt]
Few-medoids & \sndfm{$1.60 \pm 0.25$} & \best{$4.19 \pm 1.09$} & \best{$9.77 \pm 0.61$} & \best{$16.28 \pm 0.59$} & \best{$24.91 \pm 1.13$} & \best{$35.91 \pm 1.40$} & $46.49 \pm 0.82$ & $59.08 \pm 0.46$ \\
K-center    & $1.59 \pm 0.30$ & $2.38 \pm 0.45$ & $3.52 \pm 0.33$ & $6.70 \pm 0.67$ & $11.54 \pm 0.52$ & $20.01 \pm 0.42$ & $36.04 \pm 2.18$ & $55.17 \pm 1.42$ \\
Herding     & \best{$1.88 \pm 0.74$} & \sndfm{$3.27 \pm 0.76$} & \sndfm{$5.95 \pm 1.69$} & \sndfm{$11.75 \pm 2.00$} & \sndfm{$21.51 \pm 1.03$} & \sndfm{$34.80 \pm 1.25$} & \best{$50.05 \pm 1.58$} & \best{$61.80 \pm 0.96$} \\
Random      & $1.59 \pm 0.30$ & $2.17 \pm 0.95$ & $5.22 \pm 1.38$ & $9.62 \pm 1.75$ & $18.21 \pm 1.37$ & $32.52 \pm 1.54$ & \sndfm{$47.14 \pm 1.86$} & \sndfm{$59.67 \pm 1.34$} \\
PCA-guided matching    & $1.42 \pm 0.20$ & $2.40 \pm 0.53$ & $3.78 \pm 0.58$ & $7.49 \pm 0.66$ & $14.68 \pm 1.32$ & $25.49 \pm 0.70$ & $40.50 \pm 1.09$ & $57.87 \pm 1.31$ \\
\midrule
\multicolumn{9}{l}{{\it{ViT-B/16 $\to$ ViT-Small}}} \\[2pt]
Few-medoids & \best{$39.62 \pm 3.12$} & \sndfm{$51.25 \pm 1.85$} & \sndfm{$58.60 \pm 5.08$} & $66.06 \pm 1.70$ & $71.73 \pm 2.83$ & $74.02 \pm 0.80$ & $75.10 \pm 1.15$ & $77.57 \pm 0.35$ \\
K-center    & $23.50 \pm 3.44$ & $30.29 \pm 3.26$ & $39.31 \pm 3.97$ & $57.55 \pm 1.60$ & $72.67 \pm 2.30$ & $78.36 \pm 0.58$ & $81.27 \pm 0.32$ & \best{$84.33 \pm 0.25$} \\
Herding     & \sndfm{$39.08 \pm 2.32$} & \best{$51.39 \pm 3.35$} & \best{$60.99 \pm 5.37$} & \best{$74.14 \pm 1.01$} & \best{$80.21 \pm 1.28$} & \best{$82.21 \pm 0.68$} & \best{$82.66 \pm 0.35$} & \sndfm{$84.18 \pm 0.21$} \\
Random      & $23.50 \pm 3.44$ & $37.63 \pm 4.67$ & $54.62 \pm 4.28$ & \sndfm{$68.77 \pm 3.34$} & \sndfm{$78.65 \pm 0.91$} & \sndfm{$81.49 \pm 0.64$} & \sndfm{$81.77 \pm 0.45$} & $83.48 \pm 0.19$ \\
PCA-guided matching    & $16.73 \pm 1.35$ & $26.65 \pm 3.65$ & $39.75 \pm 3.55$ & $54.05 \pm 3.65$ & $68.80 \pm 2.96$ & $74.54 \pm 0.62$ & $78.09 \pm 0.47$ & $81.93 \pm 0.25$ \\
\bottomrule
\end{tabular}}
\vspace{-0.5cm}
\end{table}


\noindent
\textbf{Results.} In Table~\ref{tab:results_cifar10}, few-medoids demonstrates strong CIFAR-10 performance across two KD model pairs. For ResNet-34$\to$ResNet-18, it achieves top accuracy for five out of eight few-shot scenarios, outperforming herding and random selection by over $5\%$ in some cases. Herding marginally leads at $k=2$ and $k=4$, while random selection leads at $k=128$. For ViT-B/16$\to$ResNet-50, herding attains the best score for $k \leq 8$, but few-medoids dominates for $k \geq 16$, outperforming the other baselines by margins of up to $11\%$. 
In Table~\ref{tab:results_cifar100}, we show results on CIFAR-100, where we observe a clear $k$-dependent trend across both the ResNet-34$\to$ResNet-18 and ViT-B/16$\to$ResNet-50 settings. Few-medoids dominates the low- and mid-$k$ regimes ($k \le 16$ and $k \le 32$), outperforming herding and random selection by up to $4\%$. However, at larger coreset dimensions ($k \ge 64$), both herding and random selection overtake few-medoids.
As per Table~\ref{tab:results_flowers102}, the rankings seem unstable on Oxford Flowers 102, since the top performance alternates among few-medoids, herding, random selection, and PCA-guided matching. This sensitivity is likely due to the limited number of training images in the dataset (10 per class) and the highly imbalanced test set. 
In contrast, the results reported in Table~\ref{tab:results_food101} demonstrate that few-medoids dominates on Food-101, where it achieves the highest accuracy across all coreset dimensions for the first two KD model pairs. In the ResNet-34$\to$ResNet-18 case, it outperforms herding by up to $8\%$, and random selection by up to $10\%$, at $k=16$. This advantage extends to ViT-B/16$\to$ResNet-50, where few-medoids surpasses herding and random selection by up to $10\%$ and $11\%$ when $k=16$, respectively.

\begin{table}[t]
\caption{Few-shot KD test accuracy (\%) on Oxford Flowers 102, for $k \in \{1,2,4,8\}$. Average accuracy rates and standard deviations are computed over five runs with different seeds. \textbf{\textcolor{green!55!black}{Bold green}} indicates the best result among coreset selection methods for a given $k$ and a given neural model pair, while \textcolor{blue}{\uline{blue underline}} indicates the second-best result.}
\label{tab:results_flowers102}
\vspace{-0.3cm}
\footnotesize
\begin{tabular*}{\hsize}{@{\extracolsep{\fill}}lcccc@{}}
\toprule
Method & $k{=}1$ & $k{=}2$ & $k{=}4$ & $k{=}8$ \\
\midrule
\multicolumn{5}{l}{{\it{ResNet-34 $\to$ ResNet-18}}} \\[2pt]
Few-medoids & \sndfm{$13.91 \pm 0.74$} & \best{$25.09 \pm 0.57$} & \sndfm{$37.12 \pm 0.71$} & $51.44 \pm 0.59$ \\
K-center    & $12.86 \pm 0.60$ & $20.87 \pm 0.67$ & $33.34 \pm 0.89$ & $51.16 \pm 1.19$ \\
Herding     & \best{$14.53 \pm 0.40$} & \best{$25.09 \pm 0.97$} & \best{$38.37 \pm 0.42$} & $50.90 \pm 1.57$ \\
Random      & $12.86 \pm 0.60$ & $22.52 \pm 1.85$ & $35.73 \pm 0.69$ & \sndfm{$51.69 \pm 0.73$} \\
PCA-guided matching    & $11.71 \pm 0.74$ & $22.23 \pm 1.01$ & $36.10 \pm 0.40$ & \best{$52.13 \pm 0.88$} \\
\midrule
\multicolumn{5}{l}{{\it{ViT-B/16 $\to$ ResNet-50}}} \\[2pt]
Few-medoids & \best{$2.09 \pm 0.52$} & $9.53 \pm 3.70$ & \sndfm{$25.63 \pm 2.85$} & \best{$38.89 \pm 3.20$} \\
K-center    & $1.57 \pm 0.65$ & $8.28 \pm 4.92$ & $22.01 \pm 2.56$ & $37.00 \pm 3.46$ \\
Herding     & \sndfm{$1.90 \pm 0.69$} & \sndfm{$13.45 \pm 5.08$} & $22.00 \pm 4.56$ & $35.50 \pm 3.68$ \\
Random      & $1.57 \pm 0.65$ & $10.47 \pm 3.32$ & \best{$25.64 \pm 1.85$} & $35.15 \pm 5.34$ \\
PCA-guided matching    & $1.88 \pm 0.64$ & \best{$14.05 \pm 1.97$} & $19.82 \pm 2.28$ & \sndfm{$38.83 \pm 3.00$} \\
\midrule
\multicolumn{5}{l}{{\it{ViT-B/16 $\to$ ViT-Small}}} \\[2pt]
Few-medoids & \best{$72.61 \pm 1.59$} & $80.56 \pm 3.03$ & $89.54 \pm 1.82$ & $95.54 \pm 0.70$ \\
K-center    & $66.49 \pm 2.34$ & \sndfm{$80.61 \pm 6.22$} & \sndfm{$93.67 \pm 1.31$} & $96.62 \pm 2.41$ \\
Herding     & \sndfm{$71.63 \pm 2.25$} & \best{$83.27 \pm 4.25$} & \best{$93.81 \pm 2.36$} & \sndfm{$97.06 \pm 1.04$} \\
Random      & $66.49 \pm 2.34$ & $77.62 \pm 9.99$ & $92.15 \pm 2.68$ & \best{$97.25 \pm 0.54$} \\
PCA-guided matching    & $66.87 \pm 3.34$ & $75.51 \pm 14.31$ & $91.24 \pm 3.75$ & $96.58 \pm 1.20$ \\
\bottomrule
\end{tabular*}
\vspace{-0.5cm}
\end{table}



\begin{table}[t]
\caption{Few-shot KD test accuracy (\%) on Food-101, for $k \in \{1,2,4,8,16,32,64,128\}$. Average accuracy rates and standard deviations are computed over five runs with different seeds. \textbf{\textcolor{green!55!black}{Bold green}} indicates the best result among coreset selection methods for a given $k$ and a given neural model pair, while \textcolor{blue}{\uline{blue underline}} indicates the second-best result.}
\label{tab:results_food101}
\vspace{-0.3cm}
\resizebox{\linewidth}{!}{\footnotesize\begin{tabular}{lcccccccc}
\hline
Method & $k{=}1$ & $k{=}2$ & $k{=}4$ & $k{=}8$ & $k{=}16$ & $k{=}32$ & $k{=}64$ & $k{=}128$ \\
\midrule
\multicolumn{9}{l}{{\it{ResNet-34 $\to$ ResNet-18}}} \\[2pt]
Few-medoids & \best{$3.56 \pm 0.50$} & \best{$5.94 \pm 0.27$} & \best{$9.97 \pm 0.44$} & \best{$16.73 \pm 0.41$} & \best{$28.21 \pm 0.30$} & \best{$43.36 \pm 0.53$} & \best{$59.73 \pm 0.40$} & \best{$70.13 \pm 0.19$} \\
K-center    & $2.54 \pm 0.64$ & $2.43 \pm 0.21$ & $2.81 \pm 0.20$ & $3.72 \pm 0.16$ & $5.87 \pm 0.40$ & $12.59 \pm 0.50$ & $30.74 \pm 1.26$ & $58.37 \pm 0.18$ \\
Herding     & $2.78 \pm 0.29$ & \sndfm{$5.17 \pm 0.13$} & \sndfm{$8.16 \pm 0.14$} & $11.48 \pm 0.13$ & $20.23 \pm 0.20$ & $35.33 \pm 0.32$ & $56.17 \pm 0.33$ & $69.64 \pm 0.18$ \\
Random      & $2.54 \pm 0.64$ & $4.09 \pm 0.45$ & $6.50 \pm 0.64$ & $10.37 \pm 0.67$ & $18.19 \pm 0.60$ & $33.42 \pm 0.97$ & $55.06 \pm 0.70$ & $69.09 \pm 0.19$ \\
PCA-guided matching    & \sndfm{$2.85 \pm 0.16$} & $4.65 \pm 0.10$ & $7.39 \pm 0.15$ & \sndfm{$12.19 \pm 0.33$} & \sndfm{$22.53 \pm 0.67$} & \sndfm{$37.88 \pm 0.64$} & \sndfm{$56.96 \pm 0.14$} & \sndfm{$69.70 \pm 0.29$} \\
\midrule
\multicolumn{9}{l}{{\it{ViT-B/16 $\to$ ResNet-50}}} \\[2pt]
Few-medoids & \best{$1.45 \pm 0.14$} & \best{$2.30 \pm 0.40$} & \best{$4.13 \pm 0.57$} & \best{$8.11 \pm 2.89$} & \best{$22.35 \pm 0.66$} & \best{$36.36 \pm 0.80$} & \best{$51.07 \pm 1.66$} & \best{$65.83 \pm 2.23$} \\
K-center    & $1.18 \pm 0.19$ & $1.66 \pm 0.25$ & $1.80 \pm 0.26$ & $2.41 \pm 0.28$ & $4.03 \pm 0.74$ & $8.56 \pm 0.54$ & $20.84 \pm 0.65$ & $47.19 \pm 2.73$ \\
Herding     & \sndfm{$1.21 \pm 0.18$} & \sndfm{$2.24 \pm 0.32$} & \sndfm{$3.10 \pm 1.10$} & \sndfm{$5.80 \pm 1.46$} & \sndfm{$12.06 \pm 1.65$} & \sndfm{$25.90 \pm 1.71$} & \sndfm{$46.97 \pm 0.96$} & \sndfm{$65.26 \pm 2.03$} \\
Random      & $1.18 \pm 0.19$ & $2.13 \pm 0.15$ & $2.84 \pm 0.50$ & $4.83 \pm 0.85$ & $11.18 \pm 1.39$ & $24.45 \pm 1.32$ & $42.91 \pm 1.25$ & $63.36 \pm 1.95$ \\
PCA-guided matching    & $1.02 \pm 0.06$ & $1.76 \pm 0.33$ & $1.96 \pm 0.27$ & $3.02 \pm 0.58$ & $7.33 \pm 0.88$ & $15.27 \pm 0.86$ & $33.69 \pm 1.45$ & $56.01 \pm 1.73$ \\
\midrule
\multicolumn{9}{l}{{\it{ViT-B/16 $\to$ ViT-Small}}} \\[2pt]
Few-medoids & \best{$35.80 \pm 1.27$} & \best{$51.33 \pm 2.08$} & \sndfm{$59.60 \pm 7.30$} & \sndfm{$70.68 \pm 0.58$} & $74.92 \pm 0.57$ & $78.10 \pm 0.36$ & $80.73 \pm 0.18$ & $83.82 \pm 0.09$ \\
K-center    & $21.63 \pm 1.80$ & $25.72 \pm 5.11$ & $29.30 \pm 7.88$ & $44.77 \pm 9.35$ & $64.97 \pm 0.93$ & $74.81 \pm 0.73$ & $81.44 \pm 0.45$ & $85.47 \pm 0.16$ \\
Herding     & \sndfm{$35.10 \pm 1.89$} & \sndfm{$49.95 \pm 4.96$} & \best{$63.74 \pm 1.60$} & \best{$72.45 \pm 1.68$} & \best{$79.24 \pm 0.25$} & \best{$82.53 \pm 0.11$} & \best{$85.14 \pm 0.18$} & \best{$86.86 \pm 0.07$} \\
Random      & $21.64 \pm 1.79$ & $36.06 \pm 1.25$ & $51.67 \pm 3.67$ & $68.11 \pm 1.55$ & \sndfm{$76.56 \pm 0.79$} & \sndfm{$81.41 \pm 0.41$} & \sndfm{$84.25 \pm 0.10$} & \sndfm{$86.33 \pm 0.30$} \\
PCA-guided matching    & $13.50 \pm 0.82$ & $22.40 \pm 2.84$ & $36.36 \pm 2.00$ & $54.62 \pm 2.00$ & $69.09 \pm 0.89$ & $76.89 \pm 0.39$ & $81.52 \pm 0.26$ & $85.10 \pm 0.15$ \\
\hline
\end{tabular}}
\vspace{-0.5cm}
\end{table}

\begin{figure}[t]
\begin{subfigure}[t]{.48\linewidth}
  \centering
\includegraphics[width=1.0\linewidth]{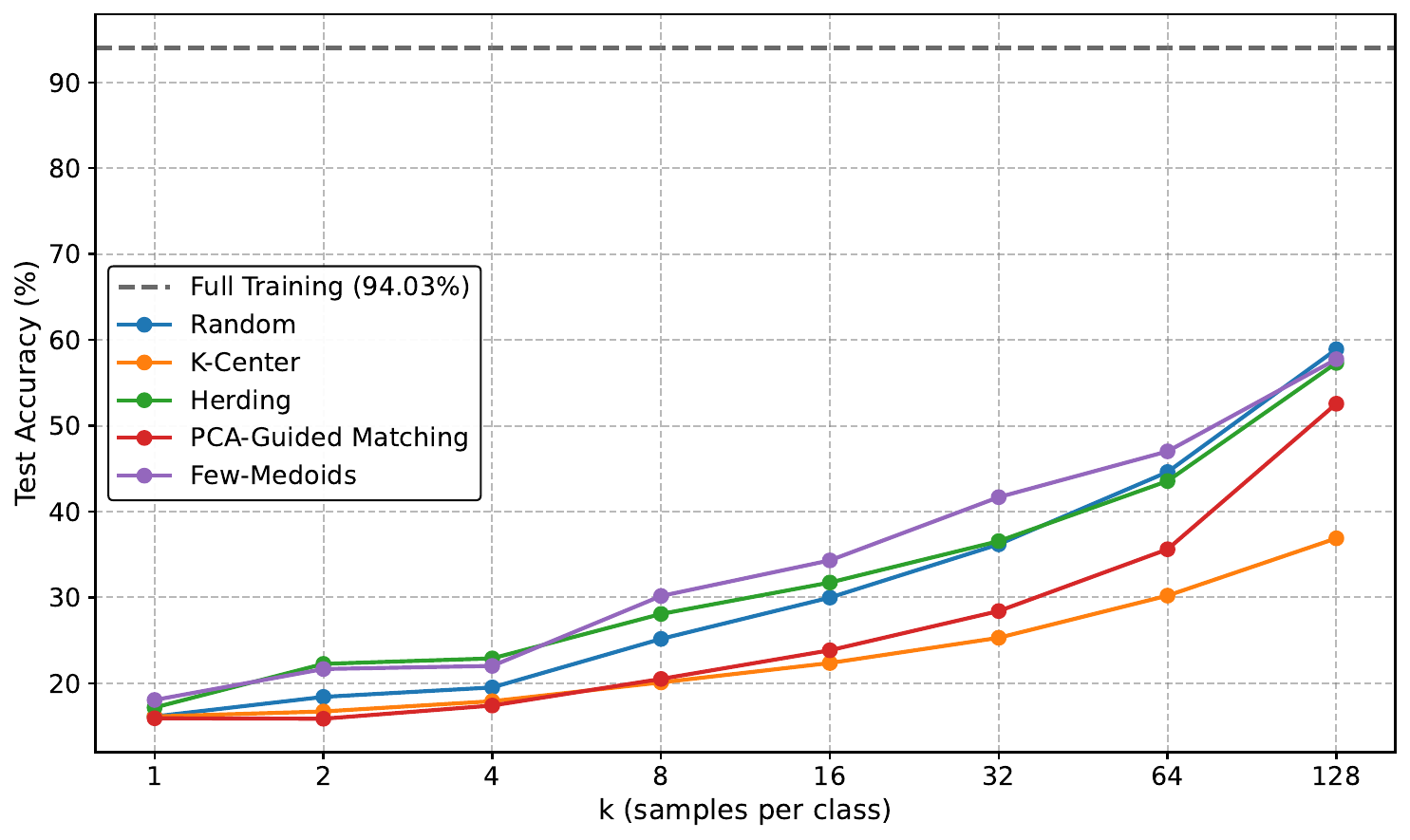}\\
  \vspace{-0.2cm}
  \caption{Few-shot KD results on CIFAR-10.}
  \label{fig:qa}
\end{subfigure}
\hfill
\begin{subfigure}[t]{.48\linewidth}
  \centering
\includegraphics[width=1.0\linewidth]{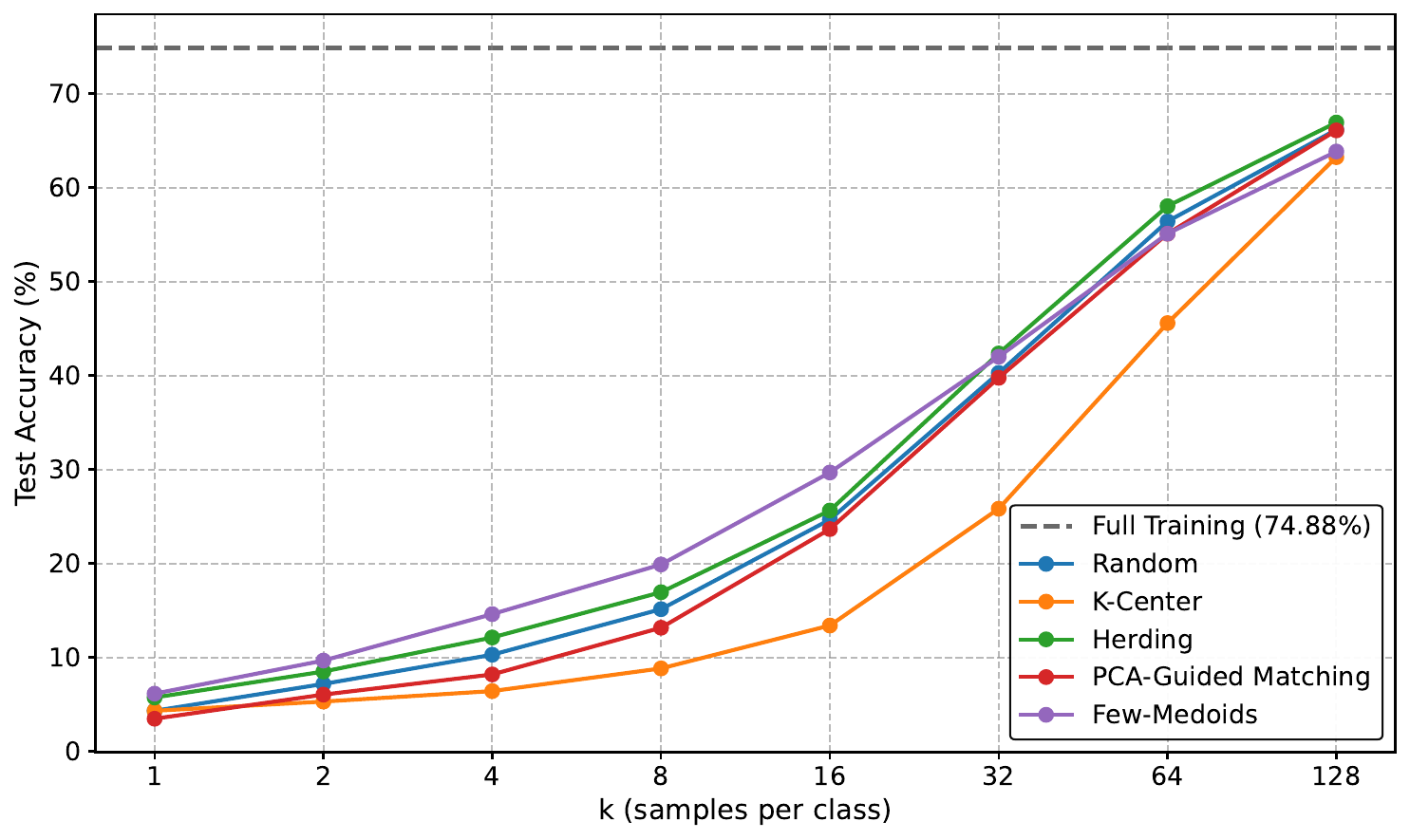}\\
\vspace{-0.2cm}
  \caption{Few-shot KD results on CIFAR-100.}
  \label{fig:qb}
\end{subfigure}
\begin{subfigure}[t]{.48\linewidth}
  \centering
\includegraphics[width=1.0\linewidth]{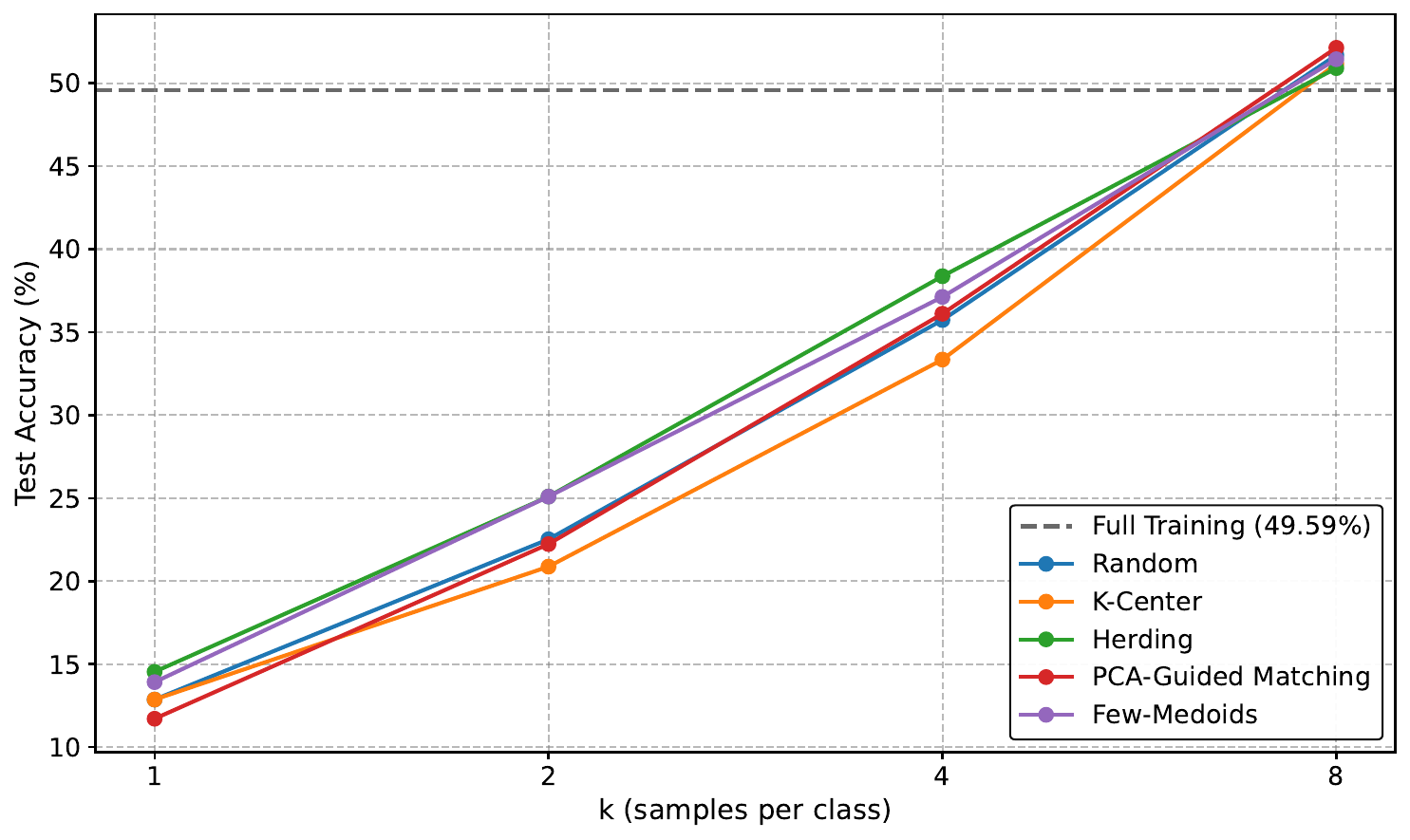}\\
  \vspace{-0.2cm}
  \caption{Few-shot KD results on Oxford Flowers 102.}
  \label{fig:qc}
    \vspace{-0.2cm}
\end{subfigure}
\hfill
\begin{subfigure}[t]{.48\linewidth}
  \centering
\includegraphics[width=1.0\linewidth]{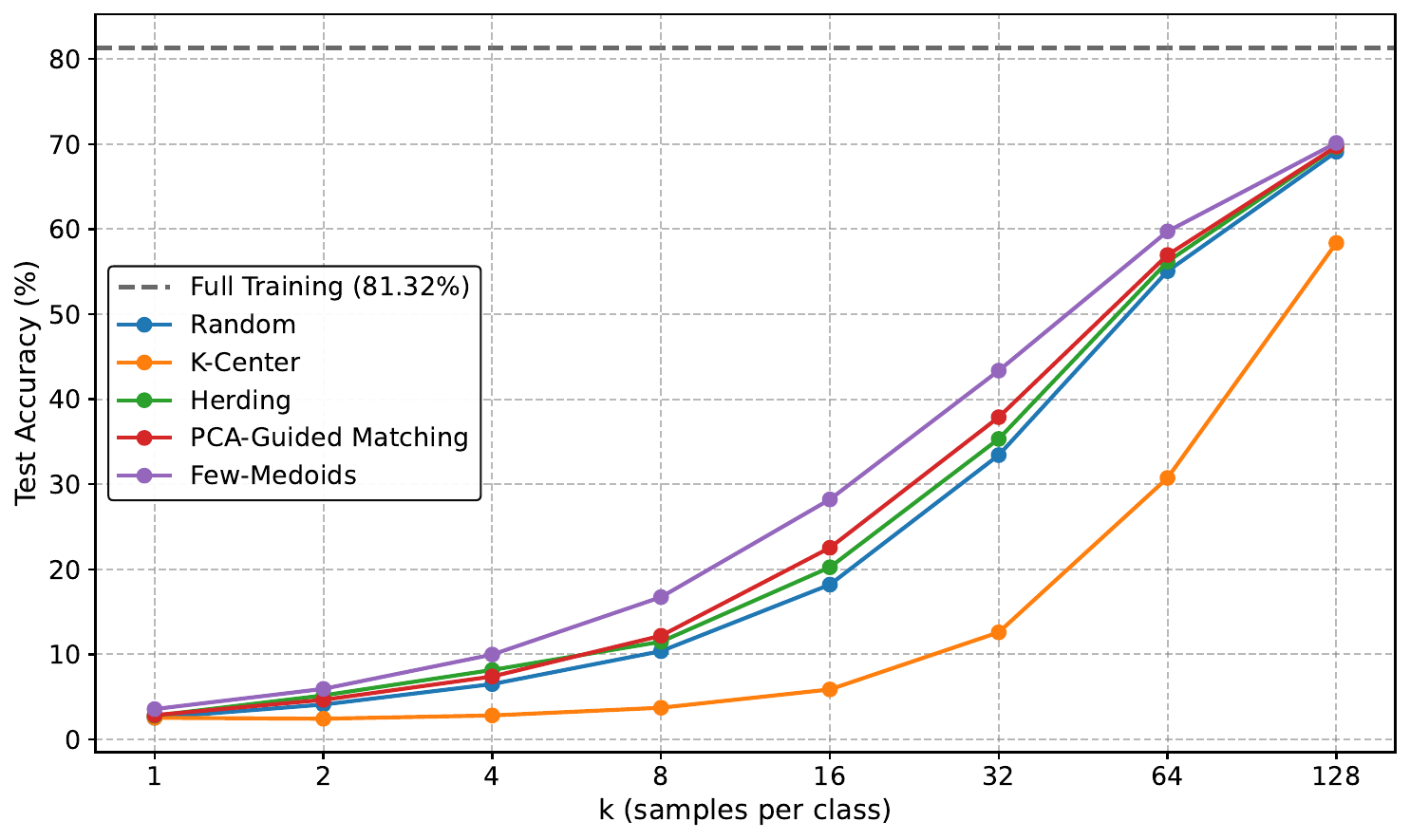}\\
\vspace{-0.2cm}
  \caption{Few-shot KD results on Food-101.}
  \label{fig:qd}
  \vspace{-0.2cm}
\end{subfigure}
\caption{Quantitative comparison for the ResNet-34$\to$ResNet-18 model pair across CIFAR-10, CIFAR-100, Oxford Flowers 102, and Food-101. Each point indicates the average accuracy over five runs. Best viewed in color.}
\label{fig:quantitative_results_resnet18}
\vspace{-0.4cm}
\end{figure}

In Figure~\ref{fig:quantitative_results_resnet18}, we illustrate the performance of ResNet-18 students for each of the studied coreset selection methods. We include these graphs as they allow a better visualization of the performance trends as a function of the coreset size, $k$. Overall, we observe that few-medoids frequently achieves the best performance, and it shows better stability across datasets compared with the other baselines.

In summary, few-medoids consistently outperforms coreset selection baselines on the first two KD model pairs (ResNet-34$\to$ResNet-18 and ViT-B/16$\to$ResNet-50) across most datasets. The main exception is the ViT-B/16$\to$ViT-Small transfer, where herding generally prevails as $k$ increases. This indicates that few-medoids excels when the student model is trained from scratch, and its initial latent space is not semantically organized a priori. For pre-trained student setups, herding seems to be more effective for $k \ge 4$, while few-medoids remains competitive for $k \le 2$.


\begin{figure}[!t]
\begin{subfigure}[t]{.48\linewidth}
  \centering
\includegraphics[width=1.0\linewidth]{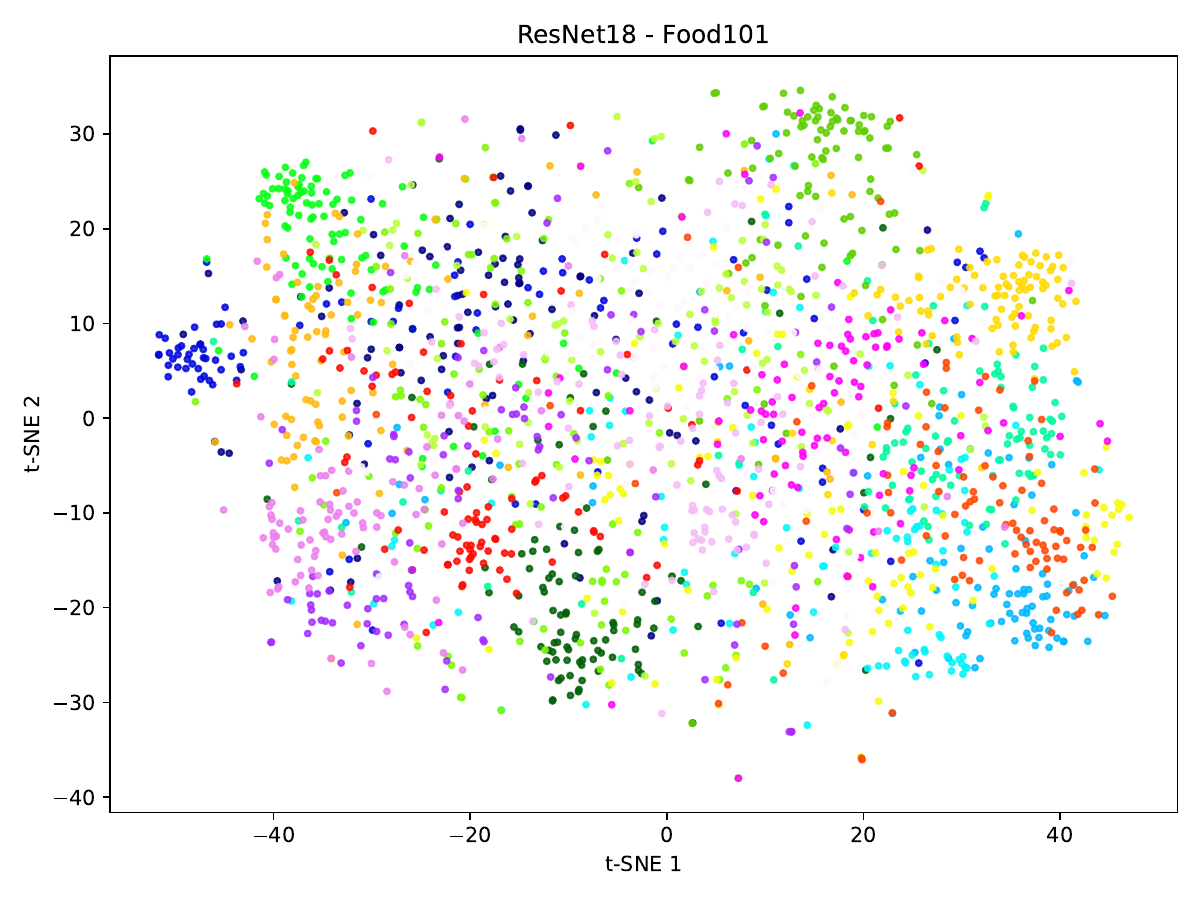}\\
  \vspace{-0.2cm}
  \caption{Random selection.}
  \label{fig:tsne_a}
    \vspace{-0.2cm}
\end{subfigure}
\hfill
\begin{subfigure}[t]{.48\linewidth}
  \centering
\includegraphics[width=1.0\linewidth]{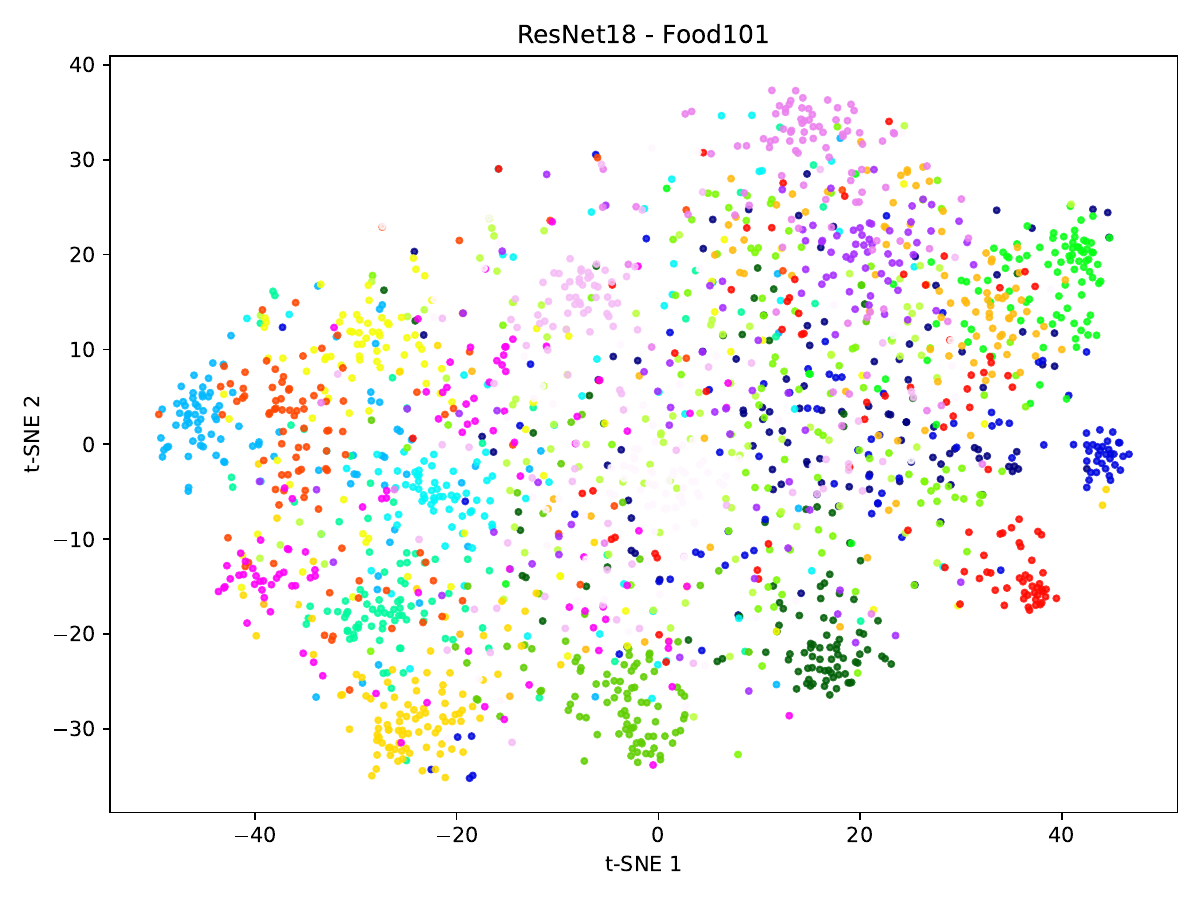}\\
\vspace{-0.2cm}
  \caption{Few-medoids.}
  \label{fig:tsne_b}
  \vspace{-0.2cm}
\end{subfigure}
\caption{t-SNE visualization of ResNet-18 embeddings ($k{=}32$) for randomly chosen test samples from 20 randomly selected classes from Food-101. Different colors represent different food categories. Few-medoids (right) generates a more discriminative latent space than random coreset selection (left). Best viewed in color.}
\label{fig:tsne}
\vspace{-0.4cm}
\end{figure}

\noindent
\textbf{Latent space analysis.} In Figure~\ref{fig:tsne}, we compare the latent space representations of ResNet-18 student models trained with few-medoids versus the random selection baseline. We can observe that few-medoids yields more well-defined class clusters than random coreset selection. Moreover, even classes that are distinguishable in both cases form tighter clusters under few-medoids, for example the \textcolor{Blue}{\textbf{dark blue}} cluster. This indicates that few-medoids learns more discriminative features, and is expected to perform better in terms of generalization capacity.

\section{Conclusion}
In this paper, we addressed the computational inefficiencies of model training by exploring the intersection of coreset selection and knowledge distillation. Moreover, we introduced few-medoids, a simple yet highly effective training-free sample selection strategy designed for the few-shot distillation framework. Rather than relying on complex heuristics or prediction difficulty, few-medoids directly leverages the latent feature space of the pre-trained teacher model. By calculating the average Euclidean distance between samples of the same class, the method selects the most geometrically central and representative samples (the class medoids) to form the coreset for student training. Our results underscored that retaining examples that are semantically relevant for the teacher, which are located near the center of the latent distribution of each class, yields a superior supervision signal for the student model, in most of the cases. Therefore, we consider that few-medoids can be employed as a strong yet simple sample selection baseline in future research on coreset selection and data pruning for knowledge distillation.

\section*{Acknowledgments}

This research is supported by the project ``Romanian Hub for Artificial Intelligence - HRIA'', Smart Growth, Digitization and Financial Instruments Program, 2021-2027, MySMIS no.~351416.

\bibliography{ref}

@techreport{Krizhevsky-2009,
  author      = {Krizhevsky, Alex and Hinton, Geoffrey},
  title       = "{Learning Multiple Layers of Features from Tiny Images}",
  institution = {University of Toronto},
  number      = {0},
  address     = {Toronto, Ontario},
  year        = {2009},
  url         = {https://www.cs.toronto.edu/~kriz/learning-features-2009-TR.pdf}
}

@inproceedings{Nilsback-ICVGIP-2008,
author = {Nilsback, Maria-Elena and Zisserman, Andrew},
title = "{Automated Flower Classification over a Large Number of Classes}",
year = {2008},
booktitle = {Proceedings of ICVGIP},
pages = {722--729},
}

@inproceedings{Bossard-ECCV-2014,
author = {Bossard, Lukas and Guillaumin, Matthieu and Van Gool, Luc},
title = "{Food-101 -- Mining Discriminative Components with Random Forests}",
year = {2014},
booktitle = {Proceedings of ECCV},
pages = {446--461},
}

@article{Hinton-arXiv-2015,
  author  = {Hinton, Geoffrey and Vinyals, Oriol and Dean, Jeff},
  title   = "{Distilling the Knowledge in a Neural Network}",
  year    = {2015},
  journal = {arXiv preprint arXiv:1503.02531},
}

@inproceedings{He-CVPR-2016,
author = {He, Kaiming and Zhang, Xiangyu and Ren, Shaoqing and Sun, Jian},
title = "{Deep Residual Learning for Image Recognition}",
year = {2016},
booktitle = {Proceedings of CVPR},
pages = {770--778},
}

@inproceedings{Dosovitskiy-ICLR-2021,
author = {Dosovitskiy, Alexey and Beyer, Lucas and Kolesnikov, Alexander and Weissenborn, Dirk and Zhai, Xiaohua and Unterthiner, Thomas and Dehghani, Mostafa and Minderer, Matthias and Heigold, Georg and Gelly, Sylvain and Uszkoreit, Jakob and Houlsby, Neil},
title = "{An Image is Worth 16x16 Words: Transformers for Image Recognition at Scale}",
year = {2021},
booktitle = {Proceedings of ICLR},
}

@inproceedings{Sener-arXiv-2017,
author = {Sener, Ozan and Savarese, Silvio},
title = {Active Learning for Convolutional Neural Networks: A Core-Set Approach},
year = {2018},
booktitle = {Proceedings of ICLR},
}

@article{Moser-arXiv-2026,
      title={A Coreset Selection of Coreset Selection Literature: Introduction and Recent Advances}, 
      author={Brian B. Moser and Arundhati S. Shanbhag and Stanislav Frolov and Federico Raue and Joachim Folz and Andreas Dengel},
      year={2026},
      journal={arXiv preprint arXiv:2505.17799},
}

@inproceedings{Welling-ICML-2009,
author = {Welling, Max},
title = {Herding Dynamical Weights to Learn},
year = {2009},
booktitle = {Proceedings of ICML},
pages = {1121--1128},
}

@inproceedings{Guo-DEXA-2022,
author = {Guo, Chengcheng and Zhao, Bo and Bai, Yanbing},
title = "{DeepCore: A Comprehensive Library for Coreset Selection in Deep Learning}",
year = {2022},
booktitle = {Proceedings of DEXA},
pages = {181--195},
}

@inproceedings{Rebuffi-CVPR-2017,
author = {Rebuffi, Sylvestre-Alvise and Kolesnikov, Alexander and Sperl, Georg and Lampert, Christoph H.},
title = "{iCaRL: Incremental Classifier and Representation Learning}",
year = {2017},
booktitle = {Proceedings of CVPR},
pages = {5533--5542},
}

@inproceedings{Chen-UAI-2010,
author = {Chen, Yutian and Welling, Max and Smola, Alex},
title = {Super-Samples from Kernel Herding},
year = {2010},
booktitle = {Proceedings of UAI},
pages = {109--116},
}

@article{Toneva-arXiv-2018,
author = {Toneva, Mariya and Sordoni, Alessandro and Tachet des Combes, Remi and Trischler, Adam and Bengio, Yoshua and Gordon, Geoffrey J.},
title = {An Empirical Study of Example Forgetting during Deep Neural Network Learning},
year = {2018},
journal = {arXiv preprint arXiv:1812.05159},
}

@inproceedings{Paul-NeurIPS-2021,
author = {Paul, Mansheej and Ganguli, Surya and Dziugaite, Gintare Karolina},
title = {Deep Learning on a Data Diet: Finding Important Examples Early in Training},
year = {2021},
booktitle = {Proceedings of NeurIPS},
pages = {20596--20607},
}

@article{Ducoffe-arXiv-2018,
author = {Ducoffe, Melanie and Precioso, Frederic},
title = {Adversarial Active Learning for Deep Networks: a Margin Based Approach},
year = {2018},
journal = {arXiv preprint arXiv:1802.09841},
}

@incollection{Agarwal-2005,
author = {Agarwal, Pankaj K. and Har-Peled, Sariel and Varadarajan, Kasturi R.},
title = {Geometric Approximation via Coresets},
year = {2005},
booktitle = {Combinatorial and Computational Geometry},
publisher = {Cambridge University Press},
pages = {1--30},
}

@inproceedings{Vaswani-NeurIPS-2017,
 author = {Vaswani, Ashish and Shazeer, Noam and Parmar, Niki and Uszkoreit, Jakob and Jones, Llion and Gomez, Aidan N and Kaiser, \L ukasz and Polosukhin, Illia},
 booktitle = {Proceedings of NeurIPS},
 title = {Attention is All you Need},
 year = {2017},
 pages={6000--6010}
}

@inproceedings{Rombach-CVPR-2022,
  title={High-resolution image synthesis with latent diffusion models},
  author={Rombach, Robin and Blattmann, Andreas and Lorenz, Dominik and Esser, Patrick and Ommer, Bj{\"o}rn},
  booktitle={Proceedings of CVPR},
  pages={10684--10695},
  year={2022}
}

@article{Touvron-arXiv-2023,
      title={LLaMA: Open and Efficient Foundation Language Models}, 
      author={Hugo Touvron and Thibaut Lavril and Gautier Izacard and Xavier Martinet and Marie-Anne Lachaux and Timothée Lacroix and Baptiste Rozière and Naman Goyal and Eric Hambro and Faisal Azhar and Aurelien Rodriguez and Armand Joulin and Edouard Grave and Guillaume Lample},
      year={2023},
      journal ={arXiv preprint arXiv 2302.13971},
}

@article{Schwartz-ACM-2020,
author = {Schwartz, Roy and Dodge, Jesse and Smith, Noah A. and Etzioni, Oren},
title = "{Green AI}",
year = {2020},
volume = {63},
number = {12},
journal = {Communications of the ACM},
pages = {54--63},
numpages = {10}
}

@article{Li-ACM-2025,
author = {Li, Pengfei and Yang, Jianyi and Islam, Mohammad A. and Ren, Shaolei},
title = "{Making AI Less `Thirsty'}",
year = {2025},
volume = {68},
number = {7},
journal = {Communications of the ACM},
pages = {54--61},
numpages = {8}
}

@inproceedings{Faiz-ICLR-2024,
title={{LLMC}arbon: Modeling the End-to-End Carbon Footprint of Large Language Models},
author={Ahmad Faiz and Sotaro Kaneda and Ruhan Wang and Rita Chukwunyere Osi and Prateek Sharma and Fan Chen and Lei Jiang},
booktitle={Proceedings of ICLR},
year={2024},
}

@InProceedings{Lee-CVPR-2024,
    author    = {Lee, Hojun and Kim, Suyoung and Lee, Junhoo and Yoo, Jaeyoung and Kwak, Nojun},
    title     = {Coreset Selection for Object Detection},
    booktitle = {Proceedings of CVPRW},
    year      = {2024},
    pages     = {7682--7691}
}

@InProceedings{Hao-CVPR-2025,
    author    = {Hao, Chenhe and Xie, Weiying and Li, Daixun and Qin, Haonan and Ye, Hangyu and Fang, Leyuan and Li, Yunsong},
    title     = "{FedCS: Coreset Selection for Federated Learning}",
    booktitle = {Proceedings of CVPR},
    year      = {2025},
    pages     = {15434--15443}
}

@inproceedings{Dai-ECCV-2020,
  title="{DA-NAS: Data Adapted Pruning for Efficient Neural Architecture Search}",
  author={Dai, Xiyang and Chen, Dongdong and Liu, Mengchen and Chen, Yinpeng and Yuan, Lu},
  booktitle={Proceedings of ECCV},
  pages={584--600},
  year={2020}
}

@inproceedings{He-CVPR-2024,
  title={Large-scale dataset pruning with dynamic uncertainty},
  author={He, Muyang and Yang, Shuo and Huang, Tiejun and Zhao, Bo},
  booktitle={Proceedings of CVPR},
  pages={7713--7722},
  year={2024}
}

@inproceedings{Baruch-ICML-2025,
title={Distilling the Knowledge in Data Pruning},
author={Emanuel Ben Baruch and Adam Botach and Igor Kviatkovsky and Manoj Aggarwal and Gerard Medioni},
booktitle={Proceedings of ICML},
year={2025},
pages={3659-3676}
}

@InProceedings{Zhao-CVPR-2022,
    author    = {Zhao, Borui and Cui, Quan and Song, Renjie and Qiu, Yiyu and Liang, Jiajun},
    title     = {Decoupled Knowledge Distillation},
    booktitle = {Proceedings of CVPR},
    year      = {2022},
    pages     = {11953--11962}
}

@article{Croitoru-TPAMI-2023,
author = {Croitoru, Florinel-Alin and Hondru, Vlad and Ionescu, Radu Tudor and Shah, Mubarak},
title = {Diffusion Models in Vision: A Survey},
year = {2023},
volume = {45},
number = {9},
journal = {IEEE Tranactions on Pattern Analysis and Machine Intelligence},
pages = {10850–10869},
}

@inproceedings{Rafailov-NeurIPS-2023,
author = {Rafailov, Rafael and Sharma, Archit and Mitchell, Eric and Ermon, Stefano and Manning, Christopher D. and Finn, Chelsea},
title = "{Direct Preference Optimization: your language model is secretly a reward model}",
year = {2023},
booktitle = {Proceedings of NeurIPS},
pages={53728--53741}
}

@INPROCEEDINGS {Liu-ICCV-2021,
author = { Liu, Ze and Lin, Yutong and Cao, Yue and Hu, Han and Wei, Yixuan and Zhang, Zheng and Lin, Stephen and Guo, Baining },
booktitle = {Proceedings of ICCV},
title = "{Swin Transformer: Hierarchical Vision Transformer using Shifted Windows}",
year = {2021},
pages = {9992--10002},
}

@inproceedings{Fukuda-Interspeech-2017,
  title={Efficient Knowledge Distillation from an Ensemble of Teachers},
  author={Takashi Fukuda and Masayuki Suzuki and Gakuto Kurata and Samuel Thomas and Jia Cui and Bhuvana Ramabhadran},
  booktitle={Proceedings of INTERSPEECH},
  year={2017},
  pages={3697--3701}
}

@inproceedings{Grigore-WACV-2025,
  title={Weight copy and low-rank adaptation for few-shot distillation of vision transformers},
  author={Grigore, Diana-Nicoleta and Georgescu, Mariana-Iuliana and Justo, Jon Alvarez and Johansen, Tor and Ionescu, Andreea Iuliana and Ionescu, Radu Tudor},
  booktitle={Proceedings of WACV},
  pages={7368--7378},
  year={2025},
}

@article{Tsuyuki-Arvix-2026,
  title="{Efficient Few-Shot Learning for Edge AI via Knowledge Distillation on MobileViT}",
  author={Tsuyuki, Shuhei and Bensaid, Reda and Morlier, J{\'e}r{\'e}my and L{\'e}onardon, Mathieu and Onizawa, Naoya and Gripon, Vincent and Hanyu, Takahiro},
  journal={arXiv preprint arXiv:2603.26145},
  year={2026}
}

@inproceedings{Sorscher-NeurIPS-2022,
  author    = {Sorscher, Ben and Geirhos, Robert and Shekhar, Shashank and Ganguli, Surya and Morcos, Ari S.},
  title     = {Beyond Neural Scaling Laws: Beating Power Law Scaling via Data Pruning},
  year      = {2022},
  booktitle = {Proceedings of NeurIPS},
  pages={19523--19536}
}

@article{Ayed-TMLR-2023,
title={Data pruning and neural scaling laws: fundamental limitations of score-based algorithms},
author={Fadhel Ayed and Soufiane Hayou},
journal={Transactions on Machine Learning Research},
issn={2835--8856},
year={2023},
}

@inproceedings{Yang-ICLR-2023,
  author    = {Yang, Shuo and Xie, Zeke and Peng, Hanyu and Xu, Min and Sun, Mingming and Li, Ping},
  title     = "{Dataset Pruning: Reducing Training Data by Examining Generalization Influence}",
  year      = {2023},
  booktitle = {Proceedings of ICLR},
}

@inproceedings{Vysogorets-ICLR-2025,
  author    = {Vysogorets, Artem and Ahuja, Kartik and Kempe, Julia},
  title     = "{DRoP: Distributionally Robust Data Pruning}",
  year      = {2025},
  booktitle = {Proceedings of ICLR},
}

@inproceedings{
Zheng-ICLR-2023,
title={Coverage-centric Coreset Selection for High Pruning Rates},
author={Haizhong Zheng and Rui Liu and Fan Lai and Atul Prakash},
booktitle={Proceedings of ICLR},
year={2023}
}

@inproceedings{Chen-ICLR-2025,
  author    = {Chen, Yudong and Xu, Xuwei and de Hoog, Frank and Liu, Jiajun and Wang, Sen},
  title     = {Medium-Difficulty Samples Constitute Smoothed Decision Boundary for Knowledge Distillation on Pruned Datasets},
  year      = {2025},
  booktitle = {Proceedings of ICLR},
}

@inproceedings{
Tan-ICLR-2025,
title={Data Pruning by Information Maximization},
author={Haoru Tan and Sitong Wu and Wei Huang and Shizhen Zhao and XIAOJUAN QI},
booktitle={Proceedings of ICLR},
year={2025}}

@inproceedings{Maharana-ICLR-2024,
  author    = {Maharana, Adyasha and Yadav, Prateek and Bansal, Mohit},
  title     = "{D2 Pruning: Message Passing for Balancing Diversity and Difficulty in Data Pruning}",
  year      = {2024},
  booktitle = {Proceedings of ICLR},
}

@inproceedings{Xia-ICLR-2023,
  author    = {Xia, Xiaobo and Liu, Jiale and Yu, Jun and Shen, Xu and Han, Bo and Liu, Tongliang},
  title     = "{Moderate Coreset: A Universal Method of Data Selection for Real-World Data-Efficient Deep Learning}",
  year      = {2023},
  booktitle = {Proceedings of ICLR},
}

@inproceedings{Everaert-ICLR-2024,
  author    = {Everaert, Dante and Potts, Christopher},
  title     = "{GIO: Gradient Information Optimization for Training Dataset Selection}",
  year      = {2024},
  booktitle = {Proceedings of ICLR},
}

@inproceedings{Mirzasoleiman-ICML-2020,
  author    = {Mirzasoleiman, Baharan and Bilmes, Jeff and Leskovec, Jure},
  title     = {Coresets for Data-efficient Training of Machine Learning Models},
  year      = {2020},
  booktitle = {Proceedings of ICML},
 pages={6950--6960}
}

@inproceedings{Killamsetty-AAAI-2021,
  author    = {Killamsetty, Krishnateja and Sivasubramanian, Durga and Ramakrishnan, Ganesh and Iyer, Rishabh},
  title     = "{GLISTER: Generalization based Data Subset Selection for Efficient and Robust Learning}",
  year      = {2021},
  booktitle = {Proceedings of AAAI},
  pages={8110--8118},
}

@inproceedings{Muennighoff-ACL-2023,
    title = "{MTEB}: Massive Text Embedding Benchmark",
    author = "Muennighoff, Niklas  and
      Tazi, Nouamane  and
      Magne, Loic  and
      Reimers, Nils",
    booktitle = "Proceedings of ACL",
    year = "2023",
    pages = "2014--2037"
}

@inproceedings{
Enevoldsen-ICLR-2025,
title={{MMTEB}: Massive Multilingual Text Embedding Benchmark},
author={Kenneth Enevoldsen and Isaac Chung and Imene Kerboua and M{\'a}rton Kardos and Ashwin Mathur and David Stap and Jay Gala and Wissam Siblini and others},
booktitle={Proceedings of ICLR},
year={2025}
}

@inproceedings{Hsu-AAAI-2015,
  title={Active learning by learning},
  author={Hsu, Wei-Ning and Lin, Hsuan-Tien},
  booktitle={Proceedings of AAAI},
  year={2015},
  pages={2659--2665}
}

@InProceedings{Parvaneh-CVPR-2022,
    author    = {Parvaneh, Amin and Abbasnejad, Ehsan and Teney, Damien and Haffari, Gholamreza (Reza) and van den Hengel, Anton and Shi, Javen Qinfeng},
    title     = {Active Learning by Feature Mixing},
    booktitle = {Proceedings of CVPR},
    year      = {2022},
    pages     = {12237--12246}
}

@InProceedings{Griffin-WACV-2026,
  author={Griffin, Brent A. and Marks, Jacob and Corso, Jason J.},
  title={Zero-Shot Coreset Selection via Iterative Subspace Sampling},
  booktitle = {Proceedings of WACV},
  year={2026},
  pages={2114--2124}
}

@inproceedings{Zheng-ICLR-2025,
title={{ELFS}: Label-Free Coreset Selection with Proxy Training Dynamics},
author={Haizhong Zheng and Elisa Tsai and Yifu Lu and Jiachen Sun and Brian R. Bartoldson and Bhavya Kailkhura and Atul Prakash},
booktitle={Proceedings of ICLR},
year={2025}
}

@article{Lupacscu-IF-2026,
  title={Large multimodal models for low-resource languages: a survey},
  author={Lupa{\c{s}}cu, Marian and Rogoz, Ana-Cristina and Stupariu, Mihai Sorin and Ionescu, Radu Tudor},
  journal={Information Fusion},
  volume={131},
  pages={104189},
  year={2026},
}

@article{Hondru-AIR-2025,
  title={Towards few-call model stealing via active self-paced knowledge distillation and diffusion-based image generation},
  author={Hondru, Vlad and Ionescu, Radu Tudor},
  journal={Artificial Intelligence Review},
  volume={58},
  number={8},
  pages={254},
  year={2025},
}
\bibliographystyle{elsarticle-harv}

\end{document}